\documentclass[sigconf,nonacm]{acmart}

\AtBeginDocument{%
  \providecommand\BibTeX{{%
    \normalfont B\kern-0.5em{\scshape i\kern-0.25em b}\kern-0.8em\TeX}}}

\setcopyright{none}

\begin{document}

\title{Energy Constraints Improve Liquid State Machine Performance}

\author{Andrew Fountain}
\email{arf8856@rit.edu}
\affiliation{
  \institution{Rochester Institute of Technology}
  \city{Rochester}
  \state{New York, USA}
}

\author{Cory Merkel}
\email{cemeec@rit.edu}
\affiliation{
  \institution{Rochester Institute of Technology}
  \city{Rochester}
  \state{New York, USA}
}


\begin{abstract}
A model of metabolic energy constraints is applied to a liquid state machine in order to analyze its effects on network performance. It was found that, in certain combinations of energy constraints, a significant increase in testing accuracy emerged; an improvement of 4.25\% was observed on a seizure detection task using a digital liquid state machine while reducing overall reservoir spiking activity by 6.9\%. The accuracy improvements appear to be linked to the energy constraints' impact on the reservoir's dynamics, as measured through metrics such as the Lyapunov exponent and the separation of the reservoir.
\end{abstract}




\maketitle

\section{Introduction}
Incorporating artificial intelligence (AI) at the edge is critical for applications which cannot support the costs of offloading computation to an external resource. For example, intelligent cameras meant to detect home intruders require a high-speed wireless connection and can have a latency of up to 5 seconds from video collection to receiving a classification result from a cloud resource, which may be untenable for some applications \cite{tang2017enabling}. However, moving AI to the edge can be difficult, as many internet enabled devices at the edge run without an operating system, which can lead to increased integration time with commonly used AI computation frameworks, such as TensorFlow. In addition, integrating AI into edge devices would also increase their computational requirements, requiring hardware acceleration or an improved processing unit, both of which may increase both the cost and the energy consumption of the device \cite{tang2017enabling}. Integrating AI in energy-constrained edge devices, such as battery-powered mobile phones, also presents difficulty as the increased computational costs would decrease the devices' battery life, which incentivises offloading of computation to external resources \cite{liang2017mobile}. In biological systems, where energy supplies are also constrained, energy constraints are leveraged to maintain stability of neuronal circuits. For example, \cite{boison2018epilepsy} suggests that disruptions in brain energy metabolism contributes to the generation and maintenance of epileptic seizures. In this work, we model the effects of metabolic energy constraints on a spiking artificial neural network and analyze their effects on network performance when undertaking computational tasks.

In the human brain, it is thought that energy metabolism is closely regulated by glial cells, the most common being the astrocyte. It is currently thought that a majority of glucose, the primary energy source for the brain, consumed in the brain is done so through glial cells. To facilitate uptake of glucose, astrocytes attach themselves to blood vessels in the brain. They then distribute energy metabolites they create to neurons that consume them for energy \cite{jha2018glia}. Astrocytes also maintain a store of glycogen, which can serve as a reserve energy supply to provide to neurons \cite{boison2018epilepsy}. This function of glial cells, where they serve to extract energy from the blood, maintain an energy store, and provide energy to neurons, is the one that will be explored in this work.

Other work has been previously performed to analyze the impact of energy constraints on artificial neural networks (ANNs). These include analyzing the impact of adding simple energy constraints to artificial spiking neurons \cite{burroni2017energetic}, creating a biorealistic neuron-glial model for energy constraints \cite{noack2017resting}, analyzing the impact of adding artificial astrocytes to a network \cite{porto2011artificial}, and also training a network to operate within arbitrarily defined energy constraints \cite{yuan2018constraints}. No work, to our knowledge, has been done to analyze the impact of energy constraints on a network trained to perform a computational task.

Previous work has also been done utilizing biologically-inspired neuron models showing good results on computational tasks, such as speech recognition. One of such works is the liquid state machine (LSM), a form of reservoir computing (RC) that makes use of spiking neurons. RC is, in essence, where a 'reservoir' of neurons or other computation units, is created with sparse, recurrent and potentially random interconnections. Inputs are fed into the reservoir, where they can spread through the reservoir over time. It has been previously shown that LSMs perform well on speech recognition \cite{jin2017performance} as well as other time-series tasks, including gait recognition and epileptic seizure detection \cite{polepalli2016digital}. In this work, we analyze the impact of adding energy constraints to an LSM and analyze their impact on the LSM's ability to solve real-world tasks. It is the hope that, by modelling the energy constraints imposed on our own brains, we will be able to gain some insight on the creation of neuromorphic hardware with low energy requirements.


The organization of the remainder of this paper is as follows: section \ref{section:background} will detail similar works and a background for the methods utilized. Section \ref{section:simulation_methods} will detail the methods used to simulate the constructed LSM, the datasets it was evaluated on, its topology, and how energy constraints were applied. Section \ref{section:results} portrays and explains the impact of applying energy constraints to the LSM. Finally, section \ref{section:conclusions} will conclude this work and contain ideas for future work.



\section{Background and Previous Work}\label{section:background}

\subsection{Liquid State Machines}

The LSM was first proposed in \cite{maass2002real} around the same time that the Echo State Network (ESN) was proposed in \cite{jaeger2001echo}. Both the ESN and LSM are reservoir computers -- in essence, they serve to take an input and abstract it in a higher dimensional space while maintaining information about the temporal ordering of input events. They perform this by feeding an input through a reservoir, a group of interconnected computational units. The difference between the ESN and LSM is that the LSM's reservoir is composed of spiking neurons, while the ESN's reservoir is composed of rate-based neurons.

An advantage of reservoir computing is that, during training, the intra-reservoir synaptic weights do not need to be modified. This allows for a quicker training time while still maintaining good performance. In the LSM's case, the only weights that are required to be trained are those between the reservoir and the readout layer, which maps states of the reservoir to the desired outputs.

According to \cite{maass2002real}, the state of the reservoir of the machine $M$ at a time $t$, $x^M(t)$ given an input to the reservoir at the same time can be represented as:

\begin{equation} \label{equ:lsm_state}
x^M(t) = (L^Mu)(t)
\end{equation}

\noindent
where $L^M$ represents the action of the neuronal circuit composing the reservoir.

Similarly, the function of the readout layer can be described as applying a readout function, $f^M$, that maps the current state at a time $t$ to the output. This can be seen as (\ref{equ:lsm_readout}), where $y(t)$ is the output of the LSM at time $t$:

\begin{equation} \label{equ:lsm_readout}
y(t) = f^M(x^M(t))
\end{equation}

\begin{figure}
  \includegraphics[width=0.85\linewidth]{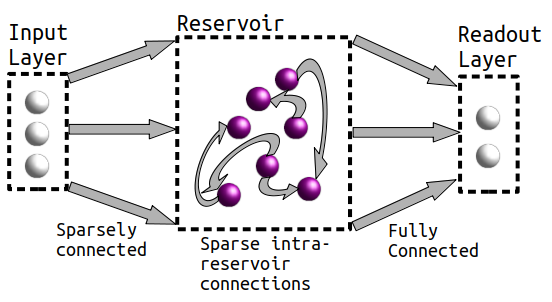}
  \caption{Diagram of an LSM. The lighter-colored spheres are rate-based neurons. While the darker-colored (purple) spheres are spiking neurons. Gray arrows are synaptic connections.}
  \label{fig:lsm}
\end{figure}

A depiction of the LSM constructed in this work can be seen as Figure \ref{fig:lsm}. Here it can be seen that the inputs to the system are sparsely connected to the reservoir, the reservoir itself is recurrently and sparsely connected and fully connected with the output layer. As per \cite{polepalli2016digital}, whose authors also constructed an LSM for their work, full connectivity with the readout layer leads to more optimal results.

Other works have employed LSMs with good results for various tasks. A very popular task for LSMs is speech recognition; other works (such as \cite{jin2017performance}) extensively analyze the performance of LSMs for speech recognition with various network parameters with generally good results. Examples of other time-series tasks LSMs have been exploited to solve include epileptic seizure detection from EEG data or user identification from their gait as measured by a smartphone accelerometer, both solved with an LSM in \cite{polepalli2016digital}.


\subsection{Energy Constrained Artificial Neural Networks}

Other works, such as \cite{burroni2017energetic} have explored the influence of energy constraints on ANNs. In the work of \cite{burroni2017energetic}, the authors give each of their spiking neurons its own energy pool, which slowly replenishes over time. Each time a neuron spikes, it drains a percentage of its energy pool. When its energy pool is depleted, the neuron is prevented from spiking. The authors evaluated the network by giving each neuron a bias and analyzing the network activity and remaining energy over time. They found that oscillations in both the spike rate and network energy were present, similar to oscillations observed in biological brain activity.

Another work by the same lab, that of \cite{noack2017resting}, developed an extension to the Izhikevich neuron model that also models the supply of energy to the neuron by a glial cell, with one glial cell per neuron. They found that, when instantiating a number of these in a network, similar spiking activity to that of a phenomenon they identified in the human brain was observed.

Work has also been done with training networks with energy constraints applied. In the work of \cite{yuan2018constraints}, the authors found that they were able to train the weights in a network such that the network operated within some specified energy constraints, with the neurons made to consume energy both from integrating inputs as well as generating spikes. They found that they were able to do this regardless of the number of neurons or synaptic connections present in the network.

In addition to work done examining the effect of energy constraints on a network, work has also been done to analyze the impact of including artificial astrocytes in a feedforward ANN \cite{porto2011artificial}. The authors of \cite{porto2011artificial} modelled the influence of astrocytes on synaptic efficacy based on neuron spiking activity. They found that, for the most part, their artificial astrocytes improved the performance of their multilayer, feedforward network \cite{porto2011artificial}.

Contrasted to this prior work performed, this present work intends to evaluate the impact of energy constraints similar to those used in the work of \cite{burroni2017energetic} on the operation of a LSM that is trained to perform a computational task. The number of neurons sharing a single energy pool is also increased past one, in an attempt to emulate the multiplicitous connection style of glial cells with neurons in mammalian brains -- according to an investigation on astrocyte connectivity within mouse brains, it was found that an astrocyte surrounds four neuron soma on average with a limit of eight \cite{halassa2007synaptic}.

\section{Simulation Methods}\label{section:simulation_methods}

\subsection{Datasets}
We evaluated our proposed LSM design using a dataset obtained through the UCI Machine Learning Repository \cite{Dua:2019} with two types of preprocessing. The task chosen is an epileptic seizure detection dataset from the work of \cite{andrzejak2001indications}. This dataset contains time-series EEG data taken from patients in five different conditions, each with 100 separate samples; measurements from healthy volunteers with their eyes open, measurements when the volunteers' eyes were closed, measurements from patients suffering from epilepsy (currently no seizure) from two different brain regions, and measurements taken during an epileptic seizure. For this work, only the first and final classes were utilized and a binary classification (seizure or not) is made by the constructed LSM. Each sample of this dataset is in the format of chunks of 4096 integers obtained when sampling at 173.61Hz. As we are only utilizing two classes of the dataset, only 200 samples were used. 70\% of these 200 samples was used for training and 30\% for testing. For the first type of preprocessing on the dataset, as recommended by the dataset collectors, a bandpass filter from 0.53 to 40 Hz was first performed on the data \cite{andrzejak2001indications}. Afterwards, feature scaling was done by dividing each timestep of each sample by the maximum of that sample. For the second preprocessing, we do the same as above but also split the time-series dataset into four channels by applying four bandpass filters, serving to perform the same preprocessing done in the work of \cite{polepalli2016digital}. The bandpass filter ranges used are 0-3Hz, 4-8Hz, 8-13Hz, and 13-30Hz \cite{polepalli2016digital}.

\subsection{Network Topology}
The topology of the LSM can be seen roughly depicted in Figure \ref{fig:lsm}; the input layer is connected sparsely to the reservoir, which is then fully connected to the output layer. The input layer of the network can be thought of as a single neuron per input feature of the dataset. The EEG seizure detection dataset, for example, only has one feature and only one input node. That input node is then wired directly to randomly selected neurons in the reservoir with a specific probability, seen as $p_{input}$ in Table \ref{tab:parameters}. All reservoir neurons are leaky integrate-and-fire (LIF) neurons. Synaptic connections between that input node and the reservoir neurons are initially randomly sampled from a uniform random distribution. The size of the reservoir for both datasets evaluated was set to 100 neurons.

The synaptic weights between the reservoir and the readout layer of the LSM were obtained through the normal equations method. The readout layer is a single neuron in both tasks evaluated. A boolean value is obtained from this readout layer by thresholding the output of this single output neuron. The threshold was taken to be the average output of the output neuron when evaluating the training portion of the data. The data that the readout layer acts on is the activity of the reservoir after the final timestep of the input has been fed into the reservoir. The reservoir activity is an array of all reservoir neuron activities. Reservoir neuron activity increases each time that neuron spikes and slowly linearly decays during timesteps where the neuron did not spike. This method of tracking neuron spiking activity was adapted from a simplified short-term plasticity method presented in \cite{soures2017deep}.

For both versions of the EEG seizure dataset, the time-series data is not converted into spike trains to reduce the computational costs of the network. Instead, each channel of each dataset is fed as direct current input to LIF neurons in the reservoir to continuously accumulate their membrane potentials, as opposed to having input spikes do the same intermittently. As inputs are directed into the reservoir as direct current inputs to the reservoir's LIF neurons, the current value for a recurrent spike was selected according to the magnitude of the inputs. This was done by taking the average magnitude of the inputs over all timesteps and all samples and multiplying it by a scalar to account for the relatively low frequency of recurrent connections compared to the inputs, that contribute once every timestep. This scalar can be seen as $\alpha_{recurrent}$ in Table \ref{tab:parameters}.

Neurons in the reservoir come in two varieties; either inhibitory or excitatory, the difference being the sign of their synaptic weights. Reservoir neurons have a 15\% chance of being inhibitory, contrasted with the commonly used value of 20\%. This difference was created after observing over-inhibition of the reservoir. All outbound synapses for an inhibitory neuron have a negative value, while those of an excitatory neuron will be positive. Each neuron within the reservoir is randomly connected with other neurons in the reservoir to implement recurrent connections. Intra-reservoir connections are made by (\ref{equ:connectivity}), which was adopted from the work of \cite{maass2002real}. The location of each neuron was chosen to be a random point in a 1x1x1 3D grid, with each dimension of the neuron's location chosen by sampling from a uniform random distribution. Parameter values used in (\ref{equ:connectivity}) can be found in Table \ref{tab:parameters}. Each intra-reservoir connection weight is initially set by sampling from a uniform random distribution. After all connections are formed, synaptic scaling is performed. This is done by, for each reservoir neuron, having the sums of all inbound synapses of a certain type equate to a certain value. The sums for different synapse types along with the other parameters used when evaluating the LSM can be found tabulated within Table \ref{tab:parameters}. All results presented are obtained when utilizing this set of parameters unless otherwise stated. We have not yet thoroughly explored the impact of these network properties on the influence of applied energy constraints and we have obtained varying results through different combinations of reservoir parameters.

\begin{equation} \label{equ:connectivity}
p_{i,j} = C_{i,j}*e^{-(\sqrt{(x_i-x_j)^2+(y_i-y_j)^2+(z_i-z_j)^2}/L_{i,j})^2}
\end{equation}

In (\ref{equ:connectivity}), $p_{i,j}$ is the probability of connection between neuron i and j in the reservoir, $x_i,y_i,z_i$ is the location of neuron i, $C_{i,j}$ is a scalar modifying how likely that connection is to occur, and $L_{i,j}$ is a scalar that controls the average length of synaptic connections. The values for both $C_{i,j}$ and $L_{i,j}$ depend on which type of neuron neurons i and j are; the different values for $C_{i,j}$ and $L_{i,j}$ can be found tabulated in Table \ref{tab:parameters}. All of the values seen in Table \ref{tab:parameters} were chosen experimentally and are likely not ideal.

\begin{table}
  \caption{Liquid State Machine Connectivity Hyperparameters}
  \label{tab:parameters}
  \begin{tabular}{ccp{0.6\linewidth}}
    \toprule
    Parameter&Value&Description\\
    \midrule
    $C_{E,E}$ & 1 & Maximum connection chance, excitatory to excitatory\\
    $C_{I,E}$ & 0.2 & Maximum connection chance, inhibitory to excitatory\\
    $C_{E,I}$ & 0.3 & Maximum connection chance, excitatory to inhibitory\\
    $C_{I,I}$ & 0.15 & Maximum connection chance, inhibitory to inhibitory\\
    $L_{E,E}$ & 1.5 & Synaptic length modifier, excitatory to excitatory\\
    $L_{I,E}$ & 4 & Synaptic length modifier, inhibitory to excitatory\\
    $L_{E,I}$ & 4 & Synaptic length modifier, excitatory to inhibitory\\
    $L_{I,I}$ & 5 & Synaptic length modifier, inhibitory to inhibitory\\
    $\Sigma_{E}$ & 4 & Synaptic scaling for excitatory synapses\\
    $\Sigma_{I}$ & 5.5 & Synaptic scaling for inhibitory synapses\\
    $\Sigma_{input}$ & 7 & Synaptic scaling for input connections\\
    $\alpha_{recurrent}$ & 5 & Recurrent synapse spike current modifier\\
    $p_{input}$ & 0.50 & Connection chance, input to reservoir neuron\\
  \bottomrule
\end{tabular}
\end{table}

In addition, in order to draw parallels to the results seen for the digital LSM presented in the work of \cite{polepalli2016digital}, we also construct a digital LSM. This was done by copying the originally constructed LSM and performing the same quantization as was done by Polepalli et al.; we quantize the input samples to 21 bits each and the intra-reservoir weights to 3 bits each. We also reduce the number of neurons in our reservoir to 60, the same used by Polepalli, et al \cite{polepalli2016digital}. We additionally quantize our measure of reservoir activity, the LIF neurons' membrane potentials, and energy, explained in the next section, to 12 bits each. Comparing to \cite{polepalli2016digital} after these changes, our architecture differs from theirs in that we do not utilize synaptic efficacy, we do not implement a multi-layer perceptron at the output layer and we also use a different metric for reservoir activity; Polepalli et al. use the number of reservoir spikes that occur in the last 1s of a sample in the dataset as a measure of reservoir activity, while we use a linearly decaying trace instead.


\subsection{Energy Constraints}

Energy constraints were added to only the reservoir of the LSM and were added in a similar manner to the method of \cite{burroni2017energetic}. Neurons were each given their own energy pool and subtracted from that pool each time the neuron fired. In addition to that, we also allow for multiple neurons to share an energy pool. In this case, the energy pool has a maximum energy linearly related to the number of neurons connected to it. For example, a neuron with a single pool was made to have a maximum energy of 1 (arbitrary) unit, and a pool with 10 neurons was made to have a maximum energy of 10 units. Note that, in our simulations, pool sizes were not mixed; each pool would have the same number of neurons attached to it.

\begin{figure}
  \includegraphics[width=0.7\linewidth]{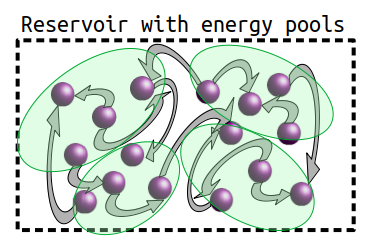}
  \caption{Diagram of the constructed reservoir with energy pools depicted. The darker-colored spheres (purple) are LIF neurons. Gray arrows are synaptic connections. The lighter-colored (green) transparent ellipses represent shared energy pools between the enclosed neurons.}
  \label{fig:energy_pooled_reservoir}
\end{figure}

In the case of multiple neurons per pool, neurons are added to the pool based on distance. To initialize, an energy pool is placed at the same location as a random neuron that does not yet have a pool. Then, the closest neurons not connected to any pool are connected to that pool until the desired number of neurons are connected to the pool. This is repeated until all reservoir neurons are connected to an energy pool. A depiction of a reservoir with 4 neurons per energy pool can be seen as Figure \ref{fig:energy_pooled_reservoir}.

Every time a reservoir neuron fires, it is required to subtract an amount of energy from its connected energy pool. This amount was varied from simulation to simulation. If the energy pool does not have enough energy, then the neuron is prohibited from firing. Each energy pool regenerates energy at a fixed rate each timestep of the simulation; this was set to 5\% of a pool's maximum energy per timestep. In addition, if an energy pool runs out of energy, the accumulated membrane potential of each attached neuron is set to its reset voltage. This is to prevent neurons from firing the instant the pool recovers due to inputs it received while the pool was disabled. Additional reasoning for this decision is that real neurons consume energy when integrating inputs, so they will not be able to integrate inputs if there is no energy to expend. This energy consumption regime leads to hysteresis behavior of a neuron's spiking activity; it will spike at some rate and then be disabled due to energy constraints. After sufficient energy regeneration, there will be a delay before spiking resumes due to the reset membrane potential.

\subsection{Evaluation Methods}
To observe the impact of energy constraints and the impact of varying energy pool sizes, a few metrics were utilized. These were training and testing accuracy on evaluated datasets along with the separation of the reservoir and the Lyapunov exponent.

\subsubsection{Separation}
The separation of a reservoir, defined in \cite{goodman2006spatiotemporal} and expanded upon in \cite{norton2010improving}, is a measure of how far applied inputs spread out in reservoir space. The expanded definition also includes intra-class variance in this metric. This metric is especially useful, as we use the normal equations method to solve for the weights between the reservoir and the output layer, which performs best if the reservoir states are linearly separable. The separation quality is defined as (\ref{equ:sep}) through (\ref{equ:individual_variance}).

The separation quality is defined as (\ref{equ:sep}) by \cite{norton2010improving} as:

\begin{equation} \label{equ:sep}
Separation = \frac{Sep_d}{Sep_v+1}
\end{equation}
\noindent
where $Sep_d$ is the inter-class distance and $Sep_v$ is the intra-class variance of the inputs abstracted in reservoir space. The inter-class distance is defined as:

\begin{equation} \label{equ:sep_d}
Sep_d = \sum_{i=1}^N\sum_{j=1}^N\frac{||C_m(O_i)-C_m(O_j)||_2}{N^2}
\end{equation}
\noindent
where $O_i$ is the set of reservoir states for input class $i$, out of N input classes. $||\cdot||_2$ denotes the L2-norm. $C_m$, the 'center of mass', is defined by \cite{goodman2006spatiotemporal} as the following:

\begin{equation} \label{equ:center_mass}
C_m(O) = \frac{\sum_{o_j \in O}o_j}{|O|}
\end{equation}
\noindent
where $|\cdot|$ denotes the cardinality of a set.\\

The intra-class variance of the separation quality is defined as:

\begin{equation} \label{equ:sep_v}
Sep_v = \frac{1}{N}*\sum_{i=1}^N\rho_i
\end{equation}
\noindent
where $\rho_i$ is computed by (\ref{equ:individual_variance}).

\begin{equation} \label{equ:individual_variance}
\rho_i = \frac{\sum_{o_j \in O_i}||C_m(O_i)-o_j||_2}{|O_i|}
\end{equation}

\subsubsection{Lyapunov Exponent}

The Lyapunov exponent is a measure of the 'chaoticness' of a system; the rate of divergence of the system's state given a small difference in inputs. An exponent greater than zero indicates that the system is chaotic; a small difference in initial inputs will result in a large change in the final outcome, while an exponent less than 0 indicates that the system is more ordered; a small difference in initial conditions will result in a small change in the final outcome. An exponent of zero is when the system is operating on the 'edge of chaos'. The Lyapunov exponent has been estimated by the authors of \cite{legenstein2007edge} by introducing a very small initial difference in reservoir state and analyzing its impact on the final reservoir state given identical inputs:

\begin{equation} \label{equ:lyapunov_maass}
\lambda \approx \frac{ln \left( \delta_{\Delta T} / \delta_o \right)}{\Delta T}
\end{equation}

In (\ref{equ:lyapunov_maass}), $\delta_o$ is the magnitude of the initial injected difference, $\delta_{\Delta T}$ is the difference between the two reservoir states after time $\Delta T$. To compute the Lyapunov exponent estimate seen as (\ref{equ:lyapunov_maass}), a single initial spike is generated in a random reservoir neuron in the network to form $\delta_o$. Then the same input is run through the LSM twice, once with the added spike and once without the added spike to obtain $\delta_{\Delta T}$. This is the method used to estimate all of the Lyapunov exponents shown in this work.

\section{Results and Analysis}\label{section:results}

The described LSM and energy constraint model were implemented and tested using a custom recurrent network simulator we built in Python. The two datasets used in this work, the 1-channel and 4-channel versions of the EEG seizure dataset, were run through the LSM while varying the energy consumption settings. Parameters varied included the number of neurons belonging to each energy pool, the amount of energy consumed from an energy pool on each reservoir neuron spike, and the random number generator seed in order to average results over many independent runs.



\begin{figure*}
  \includegraphics[width=0.85\textwidth]{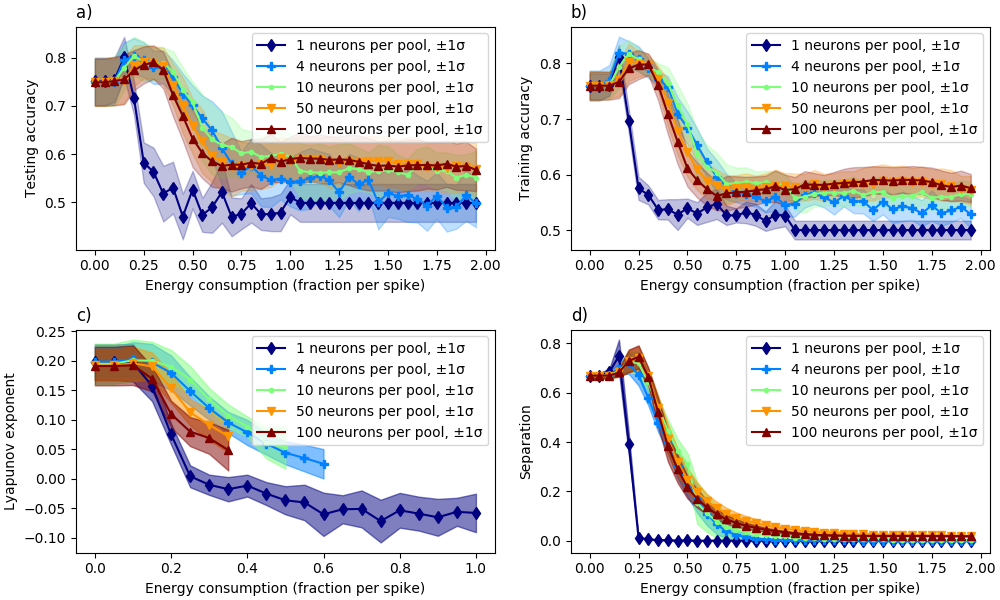}
  \caption{Plots obtained by evaluating the LSM with the EEG seizure detection dataset. The plots display energy consumption against a) average accuracy on the testing set b) average accuracy on the training set c) the average Lyapunov exponent estimate and d) the average reservoir separation. All averages were performed by collecting results over 20 independent runs. One standard deviation is shown in each plot through the shaded region surrounding each curve.}
  \label{fig:noiseless_eeg_plots}
\end{figure*}

The results obtained from evaluating the non-digitized LSM on the single-channel seizure detection dataset can be seen in Figure \ref{fig:noiseless_eeg_plots}. A rise in both testing and training accuracy is seen as energy constraints are applied, with varying results at different pool sizes. The largest accuracy gain for this dataset was 5.42\% with four neurons per pool at an energy consumption rate of 0.20 per spike. A two-sided Student's T-test was performed, yielding a p-score of $2.92\times10^{-4}$, indicating that this accuracy increase is significant. This increase in accuracy can be attributed to a rise in the separation quality of the reservoir. By definition, the separation of the reservoir is a metric of how distinguishable the different input classes are after abstraction by the reservoir. Seen in Figure \ref{fig:noiseless_eeg_plots}d is a rise in the separation of the reservoir that coincides with the increase in both testing and training accuracy on the seizure detection dataset. On investigating the influence of the energy constraints on the separation, it was found that, as energy constraints were applied both the separation distance, described by (\ref{equ:sep_d}), and the intra-class variance described by (\ref{equ:sep_v}) both decrease. However, the intra-class variance initially decayed at a higher rate than the inter-class distance. The decrease of both of these quantities was expected, as energy constraints tend to drive the activity of the reservoir to zero, but it is currently not known why the intra-class variance decreased at a higher rate, yielding a positive change in separation. 

\begin{figure*}
  \includegraphics[width=\textwidth]{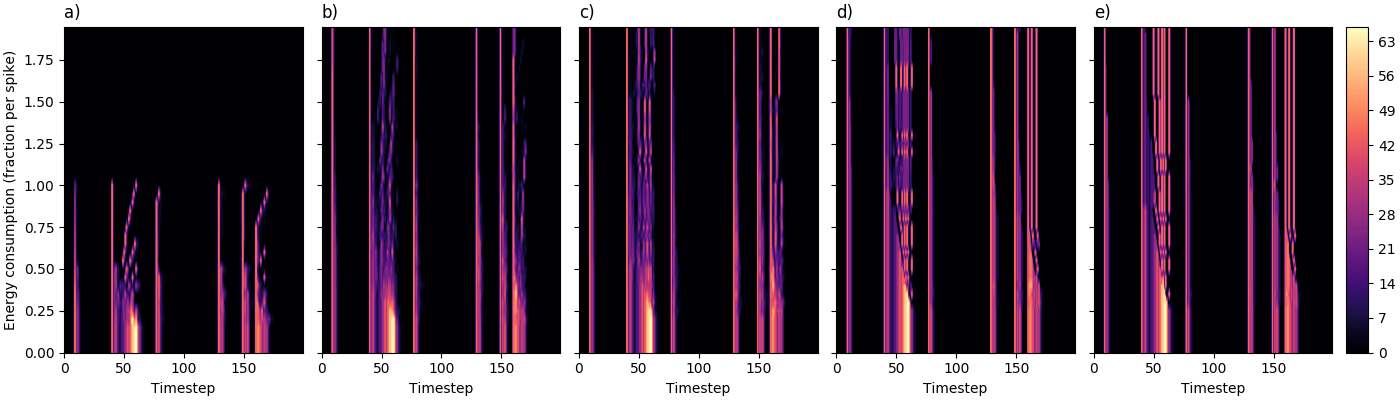}
  \caption{Total number of reservoir neuron spikes while both time and energy constraints are varied for the noiseless EEG seizure detection dataset with a) one neuron per energy pool b) four neurons per energy pool c) ten neurons per energy pool d) 50 neurons per energy pool and e) 100 neurons per energy pool. For this case, only 100 neurons were present in the reservoir. These plots depict spiking activity for a randomly selected sample from the seizure detection dataset for the first 200 timesteps.}
  \label{fig:noiseless_eeg_spikes}
\end{figure*}

The Lyapunov exponent estimate for the reservoir using the EEG seizure detection dataset can be seen as Figure \ref{fig:noiseless_eeg_plots}c. It can be seen that the Lyapunov exponent is monotonically decreasing as the applied energy constraints are made more severe. This is a result of the tendency of the energy constraints to reduce the activity of the reservoir. An example of this tendency can be seen plotted as Figure \ref{fig:noiseless_eeg_spikes}, where it can be seen that regions of high spiking activity are attenuated as energy constraints are increased, up to the point where spiking activity is completely eliminated, at least in the case where there is one neuron per pool. This attenuation of reservoir activity will tend to decrease the Lyapunov exponent. Since the Lyapunov exponent is determined by (\ref{equ:lyapunov_maass}) as the distance between two states given a small change in inputs, if the activity of both states is lowered while maintaining the same initial state difference, then the difference will decrease as well, leading to a decrease in the Lyapunov exponent as energy constraints are applied.

Also noted in the Lyapunov exponent curve is the lack of complete data. The missing datapoints correspond to conditions where the Lyapunov exponent was driven to negative infinity, that is, there was no difference between the two final reservoir states given the initial spike. This is an artifact of the method of applying energy constraints to the system. When an energy pool is depleted, all attached neurons can no longer spike and have their membrane potentials reset. As this is the case, it is more and more likely, as the number of neurons per single energy pool increases and as the energy consumption increases, that the neurons associated with the initial spike will run out of energy and have their membrane potentials reset, potentially erasing any memory of that initial spike.

\begin{figure*}
  \includegraphics[width=0.85\textwidth]{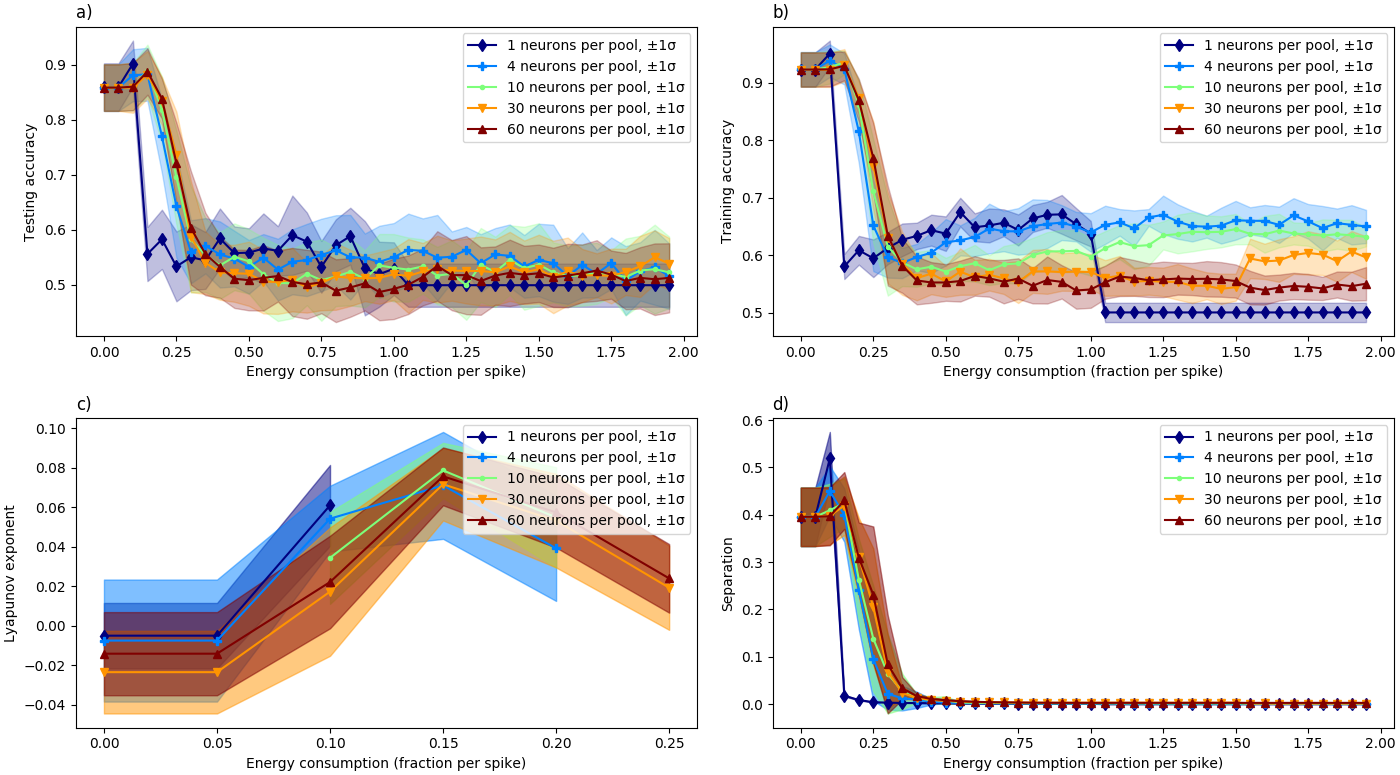}
  \caption{Plots obtained by evaluating the digital LSM on the 4-channel EEG seizure detection task. The plots display energy consumption against a) average accuracy on the testing set b) average accuracy on the training set c) the average Lyapunov exponent estimate and d) the average reservoir separation. All averages were performed by collecting results over 20 independent runs. One standard deviation is shown in each plot through the shaded region surrounding each curve.}
  \label{fig:quantized_plots}
\end{figure*}

We perform a sweep of the connectivity hyperparameters $L$ and $C$ in Table \ref{tab:parameters} for the non-digitized LSM by modifying them all by the same scalar factor. Significantly higher testing accuracy was observed with lower connection length, with the highest observed being 89.9\% at 10\% of the original connection length. A similar sweep was performed with the overall connection probability yielding similar behavior and a maximum testing accuracy of 90.4\% at 5\% of the original connection probability. Although the accuracy is much higher than with the higher connection length, the same boost in accuracy with energy constraints as seen with the less ideal connection parameters is not observed. However, despite an increase in accuracy not being present, the accuracy does not decrease significantly as energy constraints are heightened, up to a point. This leads to a reduction of reservoir spikes; without energy constraints, the reservoir was spiking just under 92800 times per 4096-timestep sample, on average. At an energy consumption of 0.15, with 50 neurons per energy pool, the reservoir spikes 87900 times, a reduction of 5\% over the original while only reducing accuracy by 0.3\%, which is well within a standard deviation of the original accuracy.

We also evaluate the digitized LSM on the 4-channel version of the EEG seizure detection task in order to compare to the results presented in \cite{polepalli2016digital}. For this task, we use the same parameters as in Table \ref{tab:parameters}, except we use an $L$ that is 0.14x as large as listed in the table, for all connection cases between excitatory and inhibitory neurons. We additionally set $\Delta t$ to a flat 0.01 instead of using the period of samples in the dataset, as was done before. This was observed to lead to better results. Results for this digital LSM can be seen as Figure \ref{fig:quantized_plots}. Without energy constraints, we obtain similar results to those of Polepalli et al., obtaining an accuracy of 85.92\%, on average, compared to their results of 85.1\% \cite{polepalli2016digital}. After adding energy constraints, we see an improvement of 4.25\% in accuracy, and the maximum accuracy observed is 90.17\%, on average, with one neuron per pool and at an energy consumption of 0.1 per spike. This increase in accuracy corresponds to a p-score of 0.0032, indicating that this result is significant. Additionally, the number of reservoir spikes is also attenuated; with no energy consumption there are just over 75400 spikes on average per 4096-timestep sample. With these energy constraints, this is lowered to just above 70200 spikes per sample, a reduction of 6.9\%. The separation of the reservoir also displays the same spike in the vicinity of the energy constraint settings that give the highest accuracy as with the non-digitized LSM results on the 1-channel EEG seizure detection task. However the Lyapunov exponent behaves significantly different; it also shows a rise to its maximum in the same place that the accuracy rises, indicating that the reservoir becomes more chaotic as energy constraints are increased, distinct from the behavior of the exponent with the 1-channel seizure detection task where it was monotonically decreasing. It is thought that this effect stems from the increased number of input features, but this is not yet known.

\section{Conclusions and Future Work}\label{section:conclusions}
The application of a model of metabolic energy constraints to a liquid state machine was seen to significantly improve testing accuracy when the LSM was applied to two versions of an EEG seizure detection dataset, with a maximum observed increase of 4.25\% to 90.17\% for a digital LSM. This increase in accuracy trended with the separation of the reservoir for both versions seizure detection dataset, lending some explanation to the phenomenon. Although it is likely that a similar accuracy improvement could be obtained by fine-tuning other hyperparameters in the LSM, varying energy constraints may be a simpler solution for, for example, a LSM implementation in hardware where modifying reservoir connectivity may not be an option. This is further supported by the observation that a boost in accuracy is seen in a digital LSM with fixed-precision weights when applying energy constraints, while also reducing the number of spikes which may lead to device power savings.

For the future, it would be interesting to implement this type of energy constrained network in hardware in order to analyze the effects of these energy constraints on device power consumption. It would also be interesting to see if a method of shaping intra-reservoir weights, which were static during this work, could be developed such that the reservoir could adapt and maintain good performance given imposed energy constraints. Another path for future work could also be examining the effects of a more realistic model of energy constraints, and their effects on other neuron behaviors, such as spike-time dependent plasticity.


\bibliographystyle{ACM-Reference-Format}
\bibliography{sample-authordraft}

\appendix

\end{document}